# A STOCHASTIC FINITE-STATE WORD-SEGMENTATION ALGORITHM FOR CHINESE

**Richard Sproat**
**Chilin Shih**
**William Gale**
*AT&T Bell Laboratories*
600 Mountain Avenue,
Room {2d-451,2d-453,2c-278}
Murray Hill, NJ, USA, 07974–0636
{rws,cls,gale}@research.att.com

**Nancy Chang**
*Harvard University*
Division of Applied Sciences
Harvard University
Cambridge, MA 02138
nchang@das.harvard.edu

cmp-lg/9405008    5 May 1994

## Abstract

We present a stochastic finite-state model for segmenting Chinese text into dictionary entries and productively derived words, and providing pronunciations for these words; the method incorporates a class-based model in its treatment of personal names. We also evaluate the system's performance, taking into account the fact that people often do not agree on a single segmentation.

## THE PROBLEM

The initial step of any text analysis task is the tokenization of the input into words. For many writing systems, using whitespace as a delimiter for words yields reasonable results. However, for Chinese and other systems where whitespace is not used to delimit words, such trivial schemes will not work. Chinese writing is *morphosyllabic* (DeFrancis, 1984), meaning that each *hanzi* – 'Chinese character' – (nearly always) represents a single syllable that is (usually) also a single morpheme. Since in Chinese, as in English, *words* may be polysyllabic, and since *hanzi* are written with no intervening spaces, it is not trivial to reconstruct which *hanzi* to group into words.

While for some applications it *may* be possible to bypass the word-segmentation problem and work straight from *hanzi*, there are several reasons why this approach will not work in a text-to-speech (TTS) system for Mandarin Chinese — the primary intended application of our segmenter. These reasons include:

1. Many *hanzi* are homographs whose pronunciation depends upon word affiliation. So, 的 is pronounced *de0*[1] when it is a prenominal modification marker, but *di4* in the word 目的 *mu4di4* 'goal'; 乾 is normally *gan1* 'dry', but *qian2* in a person's given name.

2. Some phonological rules depend upon correct word-segmentation, including Third Tone Sandhi (Shih, 1986), which changes a 3 tone into a 2 tone before another 3 tone: 小老鼠 *xiao3 [lao3 shu3]* 'lit-

   tle rat', becomes *xiao3 [ lao2-shu3 ]*, rather than *xiao2 [ lao2-shu3 ]*, because the rule first applies within the word *lao3-shu3*, blocking its phrasal application.

While a minimal requirement for building a Chinese word-segmenter is a dictionary, a dictionary is insufficient since there are several classes of words that are not generally found in dictionaries. Among these:

1. **Morphologically Derived Words**: 小將們 *xiao3-jiang4-men0* (little general-**plural**) 'little generals'.

2. **Personal Names**: 周恩來 *zhou1 en1-lai2* 'Zhou Enlai'.

3. **Transliterated Foreign Names:** 布朗士維克 *bu4-lang3-shi4-wei2-ke4* 'Brunswick'.

We present a stochastic finite-state model for segmenting Chinese text into dictionary entries and words derived via the above-mentioned productive processes; as part of the treatment of personal names, we discuss a class-based model which uses the Good-Turing method to estimate costs of previously unseen personal names. The segmenter handles the grouping of *hanzi* into words and outputs word pronunciations, with default pronunciations for *hanzi* it cannot group; we focus here primarily on the system's ability to segment text appropriately (rather than on its pronunciation abilities). We evaluate various *specific* aspects of the segmentation, and provide an evaluation of the *overall* segmentation performance: this latter evaluation compares the performance of the system with that of several human judges, since even people do not agree on a single correct way to segment a text.

## PREVIOUS WORK

There is a sizable literature on Chinese word segmentation: recent reviews include (Wang et al., 1990; Wu and Tseng, 1993). Roughly, previous work can be classified into purely statistical approaches (Sproat and Shih, 1990), statistical approaches which incorporate lexical knowledge (Fan and Tsai, 1988; Lin et al., 1993), and approaches that include lexical knowledge combined with heuristics (Chen and Liu, 1992).

---

[1] We use *pinyin* transliteration with numbers representing tones.

Chen and Liu's (1992) algorithm matches words of an input sentence against a dictionary; in cases where various parses are possible, a set of heuristics is applied to disambiguate the analyses. Various morphological rules are then applied to allow for morphologically complex words that are not in the dictionary. Precision and recall rates of over 99% are reported, but note that this covers *only* words that are in the dictionary: "the ...statistics do not count the mistakes [that occur] due to the existence of derived words or proper names" (Chen and Liu, 1992, page 105). Lin et al. (1993) describe a sophisticated model that includes a dictionary and a morphological analyzer. They also present a general statistical model for detecting 'unknown words' based on *hanzi* and part-of-speech sequences. However, their unknown word model has the disadvantage that it does not identify a sequence of *hanzi* as an unknown word *of a particular category*, but merely as an unknown word (of indeterminate category). For an application like TTS, however, it is necessary to know that a particular sequence of *hanzi* is of a particular category because, for example, that knowledge could affect the pronunciation. We therefore prefer to build particular models for different classes of unknown words, rather than building a single general model.

## DICTIONARY REPRESENTATION

The lexicon of basic words and stems is represented as a *weighted finite-state tranducer* (WFST) (Pereira et al., 1994). Most transitions represent mappings between *hanzi* and pronunciations, and are costless. Transitions between orthographic words and their parts-of-speech are represented by $\epsilon$-to-category transductions and a *unigram* cost (negative log probability) of that word estimated from a 20M *hanzi* training corpus; a portion of the WFST is given in Figure 1.[2] Besides dictionary words, the lexicon contains all *hanzi* in the Big 5 Chinese code, with their pronunciation(s), plus entries for other characters (e.g., roman letters, numerals, special symbols).

Given this dictionary representation, recognizing a *single* Chinese word involves representing the input as a finite-state acceptor (FSA) where each arc is labeled with a single *hanzi* of the input. The *left-restriction* of the dictionary WFST with the input FSA contains all and only the (single) lexical entries corresponding to the input. This WFST includes the word costs on arcs transducing $\epsilon$ to category labels. Now, input sentences consist of one or more entries from the dictionary, and we can generalize the word recognition problem to the word segmentation problem, by left-restricting the transitive closure of the dictionary with the input. The result of this left-restriction is an WFST that gives all and only the possible analyses of the input FSA into dictionary entries. In general we do not want all possible analyses but rather the *best* analysis. This is obtained by computing the least-cost path in the output WFST. The final stage of segmentation involves traversing the best path, collecting into words all sequences of *hanzi* delimited by part-of-speech-labeled arcs. Figure 2 shows an example of segmentation: the sentence 日文章魚怎麼說 "How do you say octopus in Japanese?", consists of four words, namely 日文 *ri4-wen2* 'Japanese', 章魚 *zhang1-yu2* 'octopus', 怎麼 *zen3-mo* 'how', and 說 *shuo1* 'say'. In this case, 日 *ri4* is also a word (e.g. a common abbreviation for Japan) as are 文章 *wen2-zhang1* 'essay', and 魚 *yu2* 'fish', so there is (at least) one alternate analysis to be considered.

## MORPHOLOGICAL ANALYSIS

The method just described segments dictionary words, but as noted there are several classes of words that should be handled that are not in the dictionary. One class comprises words derived by productive morphological processes, such as plural noun formation using the suffix 們 *men0*. The morphological analysis itself can be handled using well-known techniques from finite-state morphology (Koskenniemi, 1983; Antworth, 1990; Tzoukermann and Liberman, 1990; Karttunen et al., 1992; Sproat, 1992); so, we represent the fact that 們 attaches to nouns by allowing $\epsilon$-transitions from the final states of all noun entries, to the initial state of the sub-WFST representing 們. However, for our purposes it is not sufficient to represent the morphological decomposition of, say, plural nouns: we also need an estimate of the cost of the resulting word. For derived words that occur in our corpus we can estimate these costs as we would the costs for an underived dictionary entry. So, 將們 *jiang4-men0* '(military) generals' occurs and we estimate its cost at 15.02. But we also need an estimate of the probability for a non-occurring though possible plural form like 南瓜們 *nan2-gua1-men0* 'pumpkins'. Here we use the Good-Turing estimate (Baayen, 1989; Church and Gale, 1991), whereby the aggregate probability of previously unseen members of a construction is estimated as $N_1/N$, where $N$ is the total number of observed tokens and $N_1$ is the number of types observed only once. For 們 this gives $prob(unseen(們) \mid 們)$, and to get the aggregate probability of novel 們-constructions in a corpus we multiply this by $prob_{text}(們)$ to get $prob_{text}(unseen(們))$. Finally, to estimate the probability of particular unseen word 南瓜們, we use the simple bigram backoff model $prob(南瓜們) \equiv prob(南瓜)prob_{text}(unseen(們))$;

---

[2]The costs are actually for *strings* rather than *words*: we currently lack estimates for the words themselves. We assign the string cost to lexical entries with the likeliest pronunciation, and a large cost to all other entries. Thus 將/adv, with the commonest pronunciation *jiang1* has cost 5.98, whereas 將/nc, with the rarer pronunciation *jiang4*, is assigned a high cost. Note also that the current model is zeroeth order in that it uses only unigram costs. Higher order models, e.g. bigram word models, could easily be incorporated into the present architecture if desired.

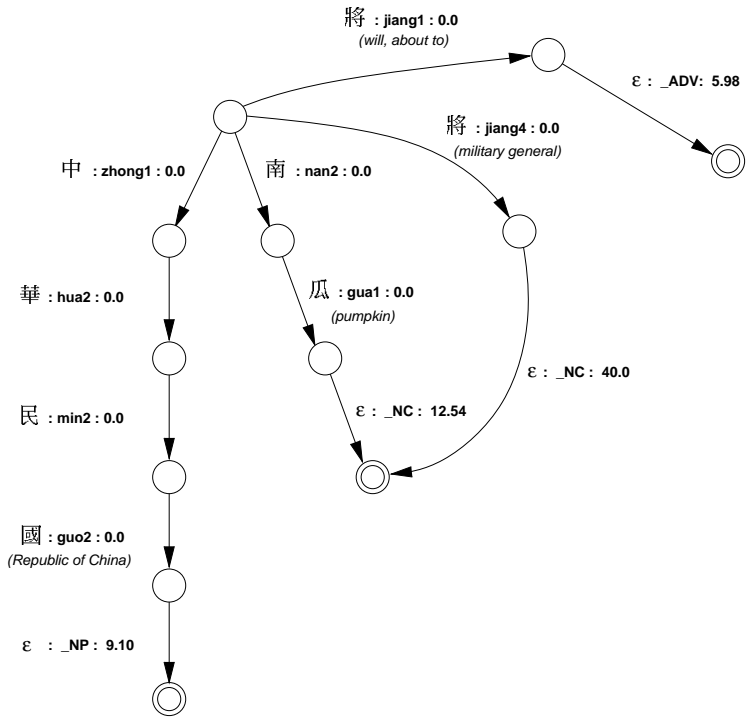

Figure 1: Partial chinese Lexicon (NC = noun; NP = proper noun)

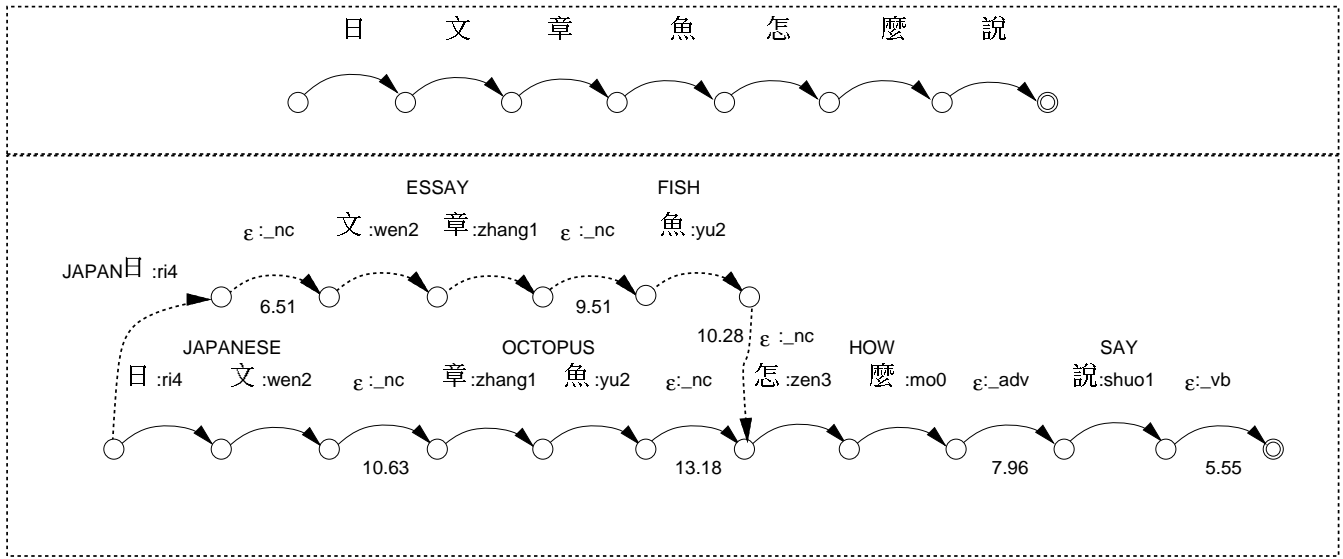

Figure 2: Input lattice (top) and two segmentations (bottom) of the sentence 'How do you say octopus in Japanese'. A non-optimal analysis is shown with dotted lines in the bottom frame.

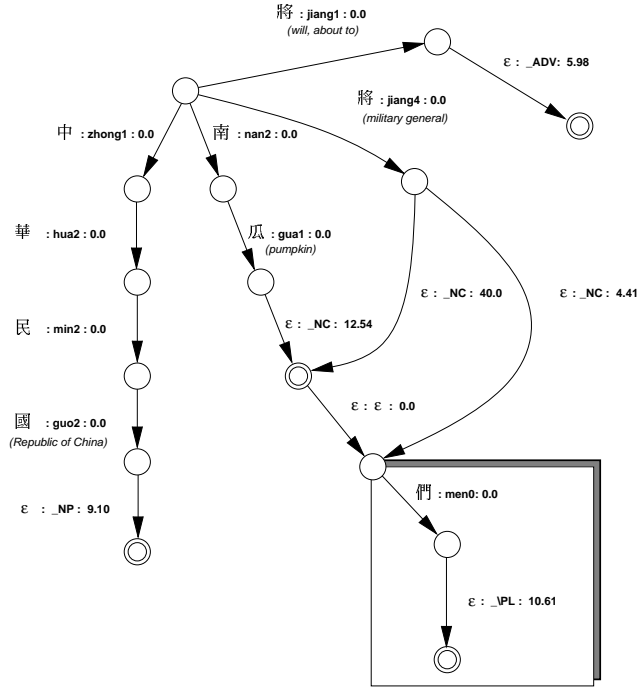

Figure 3: An example of affixation: the plural affix

$cost(南瓜們)$ is computed in the obvious way. Figure 3 shows how this model is implemented as part of the dictionary WFST. There is a (costless) transition between the NC node and 們. The transition from 們 to a final state transduces $\epsilon$ to the grammatical tag \PL with cost $cost_{text}(unseen(們))$: $cost(南瓜們) = cost(南瓜) + cost_{text}(unseen(們))$, as desired. For the seen word 將們 'generals', there is an $\epsilon$:nc transduction from 將 to the node preceding 們; this arc has cost $cost(將們) - cost_{text}(unseen(們))$, so that the cost of the whole path is the desired $cost(將們)$. This representation gives 將們 an appropriate morphological decomposition, preserving information that would be lost by simply listing 將們 as an unanalyzed form. Note that the backoff model assumes that there is a positive correlation between the frequency of a singular noun and its plural. An analysis of nouns that occur both in the singular and the plural in our database reveals that there is indeed a slight but significant positive correlation — $R^2 = 0.20$, $p < 0.005$. This suggests that the backoff model is as reasonable a model as we can use in the absence of further information about the expected cost of a plural form.

## CHINESE PERSONAL NAMES

Full Chinese personal names are in one respect simple: they are always of the form FAMILY+GIVEN. The FAMILY name set is restricted: there are a few hundred single-*hanzi* FAMILY names, and about ten double-*hanzi* ones. Given names are most commonly two *hanzi* long, occasionally one-*hanzi* long: there are thus four possible name types. The difficulty is that GIVEN names can consist, in principle, of any *hanzi* or pair of *hanzi*, so the possible GIVEN names are limited only by the total number of *hanzi*, though some *hanzi* are certainly far more likely than others. For a sequence of *hanzi* that is a *possible* name, we wish to assign a probability to that sequence *qua* name. We use an estimate derived from (Chang et al., 1992). For example, given a potential name of the form F1 G1 G2, where F1 is a legal FAMILY name and G1 and G2 are each *hanzi*, we estimate the probability of that name as the product of the probability of finding *any* name in text; the probability of F1 as a FAMILY name; the probability of the first *hanzi* of a double GIVEN name being G1; the probability of the second *hanzi* of a double GIVEN name being G2; and the probability of a name of the form SINGLE-FAMILY+DOUBLE-GIVEN. The first probability is estimated from a name count in a text database, whereas the last four probabilities are estimated from a large list of personal names.[3] This model is easily incorporated into the segmenter by building an WFST restricting the names to the four licit types, with costs on the arcs for any particular name summing to an estimate of the cost of that name. This WFST is then *summed* with the WFST implementing the dictionary and morphological rules, and the transitive closure of the resulting transducer is computed.

---

[3]We have two such lists, one containing about 17,000 full names, and another containing frequencies of *hanzi* in the various name positions, derived from a million names.

There are two weaknesses in Chang et al.'s (1992) model, which we improve upon. First, the model assumes independence between the first and second *hanzi* of a double GIVEN name. Yet, some *hanzi* are far more probable in women's names than they are in men's names, and there is a similar list of male-oriented *hanzi*: mixing *hanzi* from these two lists is generally less likely than would be predicted by the independence model. As a partial solution, for pairs of *hanzi* that cooccur sufficiently often in our namelists, we use the estimated bigram cost, rather than the independence-based cost. The second weakness is that Chang et al. (1992) assign a uniform small cost to unseen *hanzi* in GIVEN names; but we *know* that some unseen *hanzi* are merely accidentally missing, whereas others are missing for a reason — e.g., because they have a bad connotation. We can address this problem by first observing that for many *hanzi*, the general 'meaning' is indicated by its so-called 'semantic radical'. *Hanzi* that share the same 'radical', share an easily identifiable structural component: the **plant** names 蘑, 草 and 蘭 share the GRASS radical; **malady** names 癱, 痣, and 痔 share the SICKNESS radical; and **ratlike animal** names 鼠, 鼩, and 鼴 share the RAT radical. Some classes are better for names than others: in our corpora, many names are picked from the GRASS class, very few from the SICKNESS class, and none from the RAT class. We can thus better predict the probability of an unseen *hanzi* occurring in a name by computing a *within-class* Good-Turing estimate for each radical class. Assuming unseen objects *within each class* are equiprobable, their probabilities are given by the Good-Turing theorem as:

$$p_0^{cls} \propto \frac{E(N_1^{cls})}{N * E(N_0^{cls})} \quad (1)$$

where $p_0^{cls}$ is the probability of one unseen *hanzi* in class $cls$, $E(N_1^{cls})$ is the expected number of *hanzi* in $cls$ seen once, $N$ is the total number of *hanzi*, and $E(N_0^{cls})$ is the expected number of unseen *hanzi* in class $cls$. The use of the Good-Turing equation presumes suitable estimates of the unknown expectations it requires. In the denominator, the $N_0^{cls}$ are well measured by counting, and we replace the expectation by the observation. In the numerator, however, the counts of $N_1^{cls}$ are quite irregular, including several zeros (e.g. RAT, none of whose members were seen). However, there is a strong relationship between $N_1^{cls}$ and the number of *hanzi* in the class. For $E(N_1^{cls})$, then, we substitute a smooth against the number of class elements. This smooth guarantees that there are no zeroes estimated. The final estimating equation is then:

$$p_0^{cls} \propto \frac{S(N_1^{cls})}{N * N_0^{cls}} \quad (2)$$

The total of all these class estimates was about 10% off from the Turing estimate $N_1/N$ for the probability of *all* unseen *hanzi*, and we renormalized the estimates so that they would sum to $N_1/N$.

This *class-based* model gives reasonable results: for six radical classes, Table 1 gives the estimated cost for an unseen *hanzi* in the class occurring as the second *hanzi* in a double GIVEN name. Note that the good classes JADE, GOLD and GRASS have lower costs than the bad classes SICKNESS, DEATH and RAT, as desired.

## TRANSLITERATIONS OF FOREIGN WORDS

Foreign names are usually transliterated using *hanzi* whose sequential pronunciation mimics the source language pronunciation of the name. Since foreign names can be of any length, and since their original pronunciation is effectively unlimited, the identification of such names is tricky. Fortunately, there are only a few hundred *hanzi* that are particularly common in transliterations; indeed, the commonest ones, such as 巴 *ba1*, 爾 *er3*, and 阿 *a1* are often clear indicators that a sequence of *hanzi* containing them is foreign: even a name like 夏米爾 *xia4-mi3-er3* 'Shamir', which is a legal Chinese personal name, retains a foreign flavor because of 爾. As a first step towards modeling transliterated names, we have collected all *hanzi* occurring more than once in the roughly 750 foreign names in our dictionary, and we estimate the probability of occurrence of each *hanzi* in a transliteration ($p_{TN}(hanzi_i)$) using the maximum likelihood estimate. As with personal names, we also derive an estimate from text of the probability of finding a transliterated name of any kind ($p_{TN}$). Finally, we model the probability of a new transliterated name as the product of $p_{TN}$ and $p_{TN}(hanzi_i)$ for each $hanzi_i$ in the putative name.[4] The foreign name model is implemented as an WFST, which is then summed with the WFST implementing the dictionary, morphological rules, and personal names; the transitive closure of the resulting machine is then computed.

## EVALUATION

In this section we present a partial evaluation of the current system in three parts. The first is an evaluation of the system's ability to mimic humans at the task of segmenting text into word-sized units; the second evaluates the proper name identification; the third measures the performance on morphological analysis. To date we have not done a separate evaluation of foreign name recognition.

**Evaluation of the Segmentation as a Whole:** Previous reports on Chinese segmentation have invariably

---

[4]The current model is too simplistic in several respects. For instance, the common 'suffixes', *-nia* (e.g., *Virginia*) and *-sia* are normally transliterated as 尼亞 *ni2-ya3* and 西亞 *xi1-ya3*, respectively. The interdependence between 尼 or 西, and 亞 is not captured by our model, but this could easily be remedied.

Table 1: The cost as a novel GIVEN name (second position) for *hanzi* from various radical classes.

| JADE | GOLD | GRASS | SICKNESS | DEATH | RAT |
|---|---|---|---|---|---|
| 14.98 | 15.52 | 15.76 | 16.25 | 16.30 | 16.42 |

cited performance either in terms of a single percent-correct score, or else a single precision-recall pair. The problem with these styles of evaluation is that, as we shall demonstrate, even human judges do not agree perfectly on how to segment a given text. Thus, rather than give a single evaluative score, we prefer to compare the performance of our method with the judgments of *several* human subjects. To this end, we picked 100 sentences at random containing 4372 total *hanzi* from a test corpus. We asked six native speakers — three from Taiwan (T1–T3), and three from the Mainland (M1–M3) — to segment the corpus. Since we could not bias the subjects towards a particular segmentation and did not presume linguistic sophistication on their part, the instructions were simple: subjects were to mark all places they might plausibly pause if they were reading the text aloud. An examination of the subjects' bracketings confirmed that these instructions were satisfactory in yielding plausible word-sized units.

Various segmentation approaches were then compared with human performance:

1. A **greedy** algorithm, **GR**: proceed through the sentence, taking the longest match with a dictionary entry at each point.

2. An '**anti-greedy**' algorithm, **AG**: instead of the longest match, take the shortest match at each point.

3. The method being described — henceforth **ST**.

Two measures that can be used to compare judgments are:

1. **Precision**. For each pair of judges consider one judge as the standard, computing the precision of the other's judgments relative to this standard.

2. **Recall**. For each pair of judges, consider one judge as the standard, computing the recall of the other's judgments relative to this standard.

Obviously, for judges $J_1$ and $J_2$, taking $J_1$ as standard and computing the precision and recall for $J_2$ yields the same results as taking $J_2$ as the standard, and computing for $J_1$, respectively, the recall and precision. We therefore used the arithmetic mean of each interjudge precision-recall pair as a single measure of interjudge similarity. Table 2 shows these similarity measures. The average agreement among the human judges is .76, and the average agreement between **ST** and the humans is .75, or about 99% of the inter-human agreement. (**GR** is .73 or 96%.) One can better visualize the precision-recall similarity matrix by producing from that matrix a distance matrix, computing a multidimensional scaling on that distance matrix, and plotting the first two most significant dimensions. The result of this is shown in Figure 4. In addition to the automatic methods, **AG**, **GR** and **ST**, just discussed, we also added to the plot the values for the current algorithm *using only dictionary entries* (i.e., no productively derived words, or names). This is to allow for fair comparison between the statistical method, and **GR**, which is also purely dictionary-based. As can be seen, **GR** and this 'pared-down' statistical method perform quite similarly, though the statistical method is still slightly better. **AG** clearly performs much less like humans than these methods, whereas the full statistical algorithm, including morphological derivatives and names, performs most closely to humans among the automatic methods. It can be also seen clearly in this plot, two of the Taiwan speakers cluster very closely together, and the third Taiwan speaker is also close in the most significant dimension (the $x$ axis). Two of the Mainlanders also cluster close together but, interestingly, not particularly close to the Taiwan speakers; the third Mainlander is much more similar to the Taiwan speakers.

**Personal Name Identification:** To evaluate personal name identification, we randomly selected 186 sentences containing 12,000 *hanzi* from our test corpus, and segmented the text automatically, tagging personal names; note that for names there is always a single unambiguous answer, unlike the more general question of which segmentation is correct. The performance was 80.99% recall and 61.83% precision. Interestingly, Chang et al. reported 80.67% recall and 91.87% precision on an 11,000 word corpus: seemingly, our system finds as many names as their system, but with four times as many false hits. However, we have reason to doubt Chang et al.'s performance claims. Without using the same test corpus, direct comparison is obviously difficult; fortunately Chang et al. included a list of about 60 example sentence fragments that exemplified various categories of performance for their system. The performance of our system on those sentences appeared rather better than theirs. Now, on a set of 11 sentence fragments where they reported 100% recall and precision for name identification, we had 80% precision and 73% recall. However, they listed two sets, one consisting of 28 fragments and the other of 22 fragments in which they had 0% precision and recall. On the first of these our system had 86% precision and 64% recall; on the second it had 19% precision and 33% recall. Note that it is in precision that our overall performance would appear to be poorer than that of Chang et al., yet based on their published examples, our system appears to be doing better precisionwise. Thus we have some confidence that our own performance is at least as good that of (Chang et al., 1992).[5]

Table 2: Similarity matrix for segmentation judgments

| Judges | AG | GR | ST | M1 | M2 | M3 | T1 | T2 | T3 |
|---|---|---|---|---|---|---|---|---|---|
| AG | | 0.70 | 0.70 | 0.43 | 0.42 | 0.60 | 0.60 | 0.62 | 0.59 |
| GR | | | 0.99 | 0.62 | 0.64 | 0.79 | 0.82 | 0.81 | 0.72 |
| ST | | | | 0.64 | 0.67 | 0.80 | 0.84 | 0.82 | 0.74 |
| M1 | | | | | 0.77 | 0.69 | 0.71 | 0.69 | 0.70 |
| M2 | | | | | | 0.72 | 0.73 | 0.71 | 0.70 |
| M3 | | | | | | | 0.89 | 0.87 | 0.80 |
| T1 | | | | | | | | 0.88 | 0.82 |
| T2 | | | | | | | | | 0.78 |

**Evaluation of Morphological Analysis:** In Table 3 we present results from small test corpora for some productive affixes; as with names, the segmentation of morphologically derived words is generally either right or wrong. The first four affixes are so-called resultative affixes: they denote some property of the resultant state of an verb, as in 忘不了 *wang4-bu4-liao3* (forget-not-attain) 'cannot forget'. The last affix is the nominal plural. Note that 了 in 忘不了 is normally pronounced as *le0*, but when part of a resultative it is *liao3*. In the table are the (typical) classes of words to which the affix attaches, the number found in the test corpus by the method, the number correct (with a precision measure), and the number missed (with a recall measure).

## CONCLUSIONS

In this paper we have shown that good performance can be achieved on Chinese word segmentation by using probabilistic methods incorporated into a uniform stochastic finite-state model. We believe that the approach reported here compares favorably with other reported approaches, though obviously it is impossible to make meaningful comparisons in the absence of uniform test databases for Chinese segmentation. Perhaps the single most important difference between our work and previous work is the form of the evaluation. As we have observed there is often no single right answer to word segmentation in Chinese. Therefore, claims to the effect that a particular algorithm gets 99% accuracy are meaningless without a clear definition of accuracy.

## ACKNOWLEDGEMENTS

We thank United Informatics for providing us with our corpus of Chinese text, and BDC for the 'Behavior Chinese-English Electronic Dictionary'. We further thank Dr. J.-S. Chang of Tsinghua University, for kindly providing us with the name corpora. Finally, we thank two anonymous ACL reviewers for comments.

---

[5]We were recently pointed to (Wang et al., 1992), which we had unfortunately missed in our previous literature search. We hope to compare our method with that of Wang et al. in a future version of this paper.

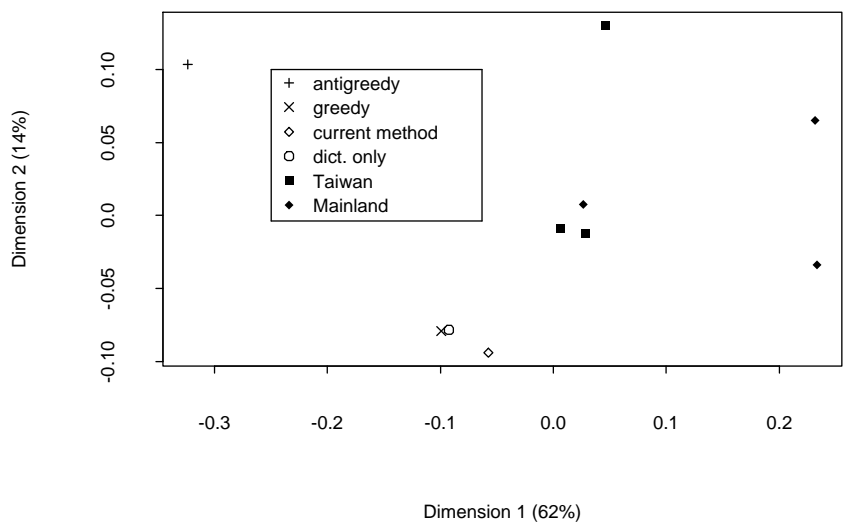

Figure 4: Classical metric multidimensional scaling of distance matrix, showing the two most significant dimensions. The percentage scores on the axis labels represent the amount of data explained by the dimension in question.

Table 3: Performance on morphological analysis.

| Affix | Pron | Base category | $N$ found | $N$ correct (prec.) | $N$ missed (rec.) |
|---|---|---|---|---|---|
| 不下 | bu2-xia4 | verb | 20 | 20 (100%) | 12 (63%) |
| 不下去 | bu2-xia4-qu4 | verb | 30 | 29 (97%) | 1 (97%) |
| 不了 | bu4-liao3 | verb | 72 | 72 (100%) | 15 (83%) |
| 得了 | de2-liao3 | verb | 36 | 36 (100%) | 11 (77%) |
| 們 | men0 | noun | 141 | 139 (99%) | 6 (96%) |